\documentclass{article}

\usepackage{microtype}
\usepackage{graphicx}
\usepackage{subfigure}
\usepackage{booktabs} 
\usepackage{amsmath}
\usepackage{mathrsfs}
\usepackage{dsfont}
\usepackage{amssymb}
\usepackage{xspace}
\usepackage{enumitem}
\usepackage[usenames,dvipsnames,svgnames,table]{xcolor}

\usepackage{hyperref}

\DeclareMathOperator*{\argmax}{arg\,max}
\DeclareMathOperator*{\argmin}{arg\,min}

\newcommand{\ie}{\textit{i}.\textit{e}.}
\newcommand{\eg}{\textit{e}.\textit{g}.}

\usepackage[accepted]{icml2019}

\icmltitlerunning{Neural Collaborative Subspace Clustering}

\begin{document}

\twocolumn[
\icmltitle{Neural Collaborative Subspace Clustering}



\icmlsetsymbol{equal}{*}

\begin{icmlauthorlist}
\icmlauthor{Tong Zhang}{anu,motovis}
\icmlauthor{Pan Ji}{nec}
\icmlauthor{Mehrtash Harandi}{monash}
\icmlauthor{Wenbing Huang}{tencent}
\icmlauthor{Hongdong Li}{anu}
\end{icmlauthorlist}

\icmlaffiliation{anu}{Australian National University}
\icmlaffiliation{nec}{NEC Labs America}
\icmlaffiliation{monash}{Monash University}
\icmlaffiliation{tencent}{Tencent AI Lab}
\icmlaffiliation{motovis}{Motovis Intelligent Technologies}

\icmlcorrespondingauthor{Pan Ji}{peterji1990@gmail.com}

\icmlkeywords{Machine Learning, ICML}

\vskip 0.3in
]



\printAffiliationsAndNotice{}  

\begin{abstract}
We introduce the Neural Collaborative Subspace Clustering, a neural model that discovers clusters of data points drawn from a union of low-dimensional subspaces. 
In contrast to previous attempts, our model runs without the aid of spectral clustering. This makes our algorithm one of the kinds that can gracefully scale to large datasets.
At its heart, our neural model benefits from a classifier which determines whether a pair of points lies on the same subspace or not.
Essential to our model is the construction of two affinity matrices, one from the classifier and the other from a notion of subspace self-expressiveness, to supervise training in a collaborative scheme. We thoroughly assess and contrast the performance of our model against various state-of-the-art  clustering algorithms including deep subspace-based ones. 

\end{abstract}
\section{Introduction}
In this paper, we tackle the problem of subspace clustering, where we aim to cluster data points drawn from a union of low-dimensional subspaces in an unsupervised manner. Subspace Clustering (SC) has achieved great success in various computer vision tasks, such as
 motion segmentation~\cite{kanatani2001motion,elhamifar2009sparse,ji2014robust,ji2016robust}, face clustering~\cite{ho2003clustering,elhamifar2013sparse} and image segmentation~\cite{yang2008unsupervised,ma2007segmentation}. 
 
Majority of the SC algorithms~\cite{yan2006general,chen2009spectral,elhamifar2013sparse,liu2013robust, wang2013provable,lu2012robust,ji2015shape,you2016oracle} rely on the linear subspace assumption to construct the affinity matrix for spectral clustering. However, in many cases data do not naturally conform to linear  models, which in turns results in the development of non-linear SC techniques. 
Kernel methods~\cite{chen2009kernel,patel2013latent, patel2014kernel,yin2016kernel,xiao2016robust,Ji2017AdaptiveLK} can be employed to implicitly map data to 
higher dimensional spaces, hoping that data conform better to linear models in the resulting spaces. 
However and aside from the difficulties of choosing the right kernel function (and its parameters), there is no theoretical guarantee 
that such a kernel exists. The use of deep neural networks as non-linear mapping functions to determine subspace friendly latent spaces has formed the latest developments in the field with  promising results~\cite{ji2017deep,peng2016deep}.

Despite significant improvements, SC algorithms still resort to spectral clustering which in hindsight requires  constructing an affinity matrix.
This step, albeit effective, hampers the scalability as it takes $O(n^2)$ memory and $O(kn^2)$ computation for storing and decomposing the affinity matrix for $n$ data points and $k$ clusters. There are several attempts to resolve the scalability issue. For example, \cite{you2016oracle} accelerate the construction of the affinity matrix using orthogonal matching pursuit; \cite{DBLP:journals/corr/abs-1811-01045} resort to $k$-subspace clustering to avoid generating the affinity matrix. However, the scalability issue either remains due to the use of spectral clustering~\cite{you2016oracle}, or mitigates but at the cost of performance~\cite{DBLP:journals/corr/abs-1811-01045}.
 
In this paper, we propose a neural structure to improve the performance of subspace clustering while being mindful to the scalablity issue. To this end, we first formulate subspace clustering as a classification problem, which in turn removes the  spectral clustering step from the computations.  
Our neural model is comprised of two modules, one for classification and one for affinity learning. Both modules collaborate during learning, hence the name ``Neural Collaborative Subspace Clustering''. During training and in each iteration, we use the affinity matrix generated by the subspace self-expressiveness to supervise the affinity matrix computed from the classification part. Concurrently, we make use of the classification part to improve self-expressiveness to build a better affinity matrix through collaborative optimization.

We evaluate our algorithm on three datasets , namely
MNIST~\cite{lecun1998gradient}, Fashion-MNIST~\cite{xiao2017/online}, and hardest one Stanford Online Products datasets~\cite{oh2016deep} which exhibit different levels of difficulty.
Our empirical study  shows the superiority of the proposed algorithm over several state-of-the-art baselines including deep subspace clustering techniques.

\section{Related Work}

\begin{figure*}[ht]
\begin{center}
\centerline{\includegraphics[width=1.7\columnwidth]{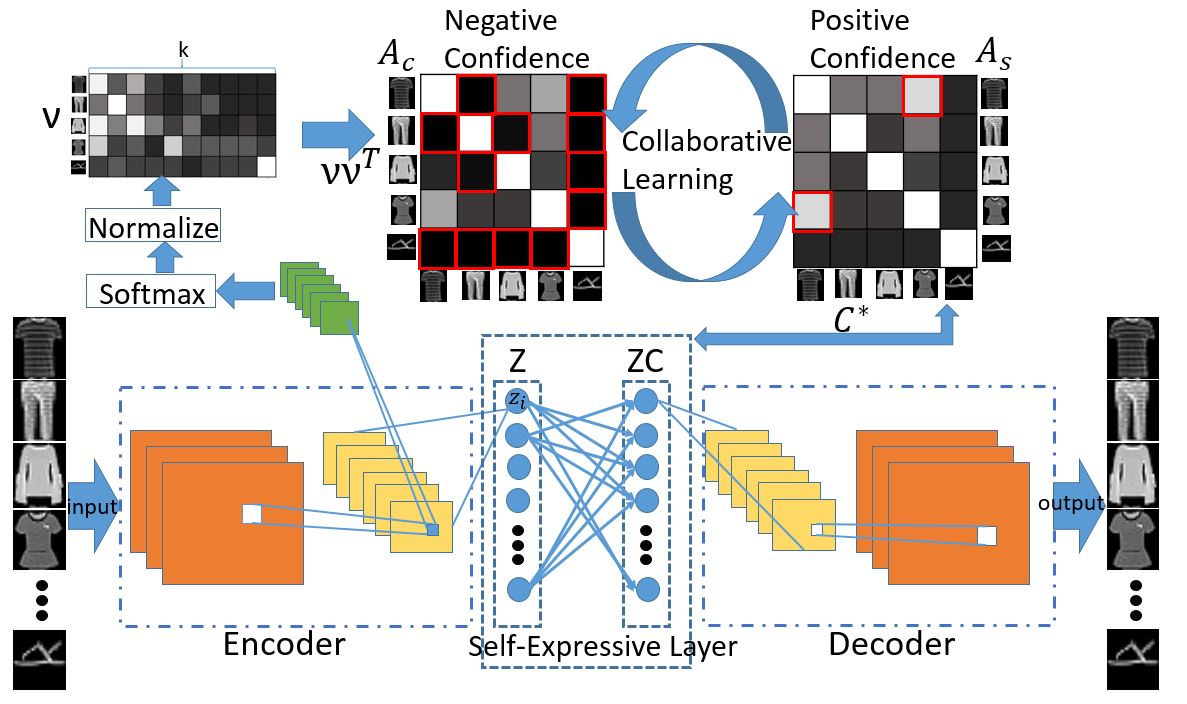}}
\caption{The Neural Collaborative Subspace Clustering framework. The affinity matrix generated by self-expressive layer, $\mathbf{A}_{s}$, and classifier, $\mathbf{A}_{c}$, supervise each other by selecting the high confidence parts for training. Red squares in $\mathbf{A}_{s}$ highlight positive pairs (belonging to the same subspace). Conversely, red squares in $\mathbf{A}_{c}$ highlight the negative pairs (belonging to different subspace). Affinities are coded with shades, meaning that light gray denotes large affinity while dark shades representing small affinities.}\label{structure}

\end{center}
\vskip -0.3in
\end{figure*}

\subsection{Subspace Clustering} 
Linear subspace clustering encompasses a vast set of techniques, 
among them, spectral clustering algorithms are more favored to cluster  high-dimensional data~\cite{vidal2011subspace}. One of the crucial challenges in employing spectral clustering on subspaces is the construction of an appropriate affinity matrix. We can categorize the 
algorithms based on the way the affinity matrix is constructed into three main groups: factorization based methods~\cite{gruber2004multibody,mo2012semi}, model based methods~\cite{chen2009spectral, ochs2014segmentation, purkait2014clustering}, self-expressiveness based methods~\cite{elhamifar2009sparse, ji2014efficient, liu2013robust, vidal2014low}. 

The latter, \ie, self-expressiveness based methods have become dominant due to their elegant convex formulations and existence of theoretical analysis. The basic idea of subspace self-expressiveness is that one point can be represented in terms of a linear combination of other points from the same subspace. This leads to several advantages over other methods: (i) it is more robust to noise and outliers; (ii) the computational complexity of the self-expressiveness affinity  does not grow exponentially with the number of subspaces and their dimensions; (iii) it also exploits the non-local information without the need of specifying the size of the neighborhood (\ie, the number of nearest neighbors as usually 
used for identifying locally linear subspaces~\cite{yan2006general,zhang2012hybrid}).

The assumption of having linear subspaces does not necessarily hold in practical problems. Several works are proposed to tackle the situation where data points do not form linear subspaces but nonlinear ones. Kernel sparse subspace clustering (KSSC)~\cite{patel2014kernel} and Kernel Low-rank representation~\cite{xiao2016robust} benefit from pre-defined kernel functions, such as polynomial or Radial Basis Functions (RBF), to cast the problem in high-dimensional (possibly infinite) reproducing kernel Hilbert spaces. However, it is still not clear how to choose proper kernel functions for different datasets and there is no guarantee that the feature spaces generated by kernel tricks are well-suited to linear subspace clustering. 

Recently, Deep Subspace Clustering Networks (DSC-Net)~\cite{ji2017deep} are introduced to tackle the non-linearity arising in subspace clustering, where data is non-linearly mapped to a latent space with convolutional auto-encoders and a new {\it self-expressive layer} is introduced between the encoder and decoder to facilitate an end-to-end learning of the affinity matrix. 
Although DSC-Net outperforms traditional subspace clustering methods by large, their 
computational cost and memory footprint can become overwhelming even for mid-size problems.

There are a few attempts to tackle the scalability of subspace clustering. The SSC-Orthogonal Matching Pursuit (SSC-OMP)~\cite{you2016scalable} replaces the large scale convex optimization procedure with the OMP algorithm to represent the affinity matrix. However, SSC-OMP sacrifices the clustering performance in favor of speeding up the computations, and it still may fail when the number of data points is very large. 
$k$-Subspace Clustering Networks ($k$-SCN)~\cite{DBLP:journals/corr/abs-1811-01045} is proposed to make subspace clustering applicable to large datasets. This is achieved via bypassing the construction of affinity matrix and consequently avoiding spectral clustering, and introducing the iterative method of $k$-subspace clustering~\cite{tseng2000nearest,bradley2000k} into a deep structure. Although $k$-SCN develops two approaches to update the subspace and networks, it still shares the same drawbacks as iterative methods, for instance, it requires a good initialization, and seems fragile to outliers.

\subsection{Model fitting}  In learning theory, distinguishing outliers and noisy samples from clean ones to facilitate training is an active research topic. For example, Random Sample Consensus (RANSAC)~\cite{fischler1981random} is a classical and well-received algorithm for fitting a model to a cloud of points corrupted by noise. Employing RANSAC on subspaces~\cite{yang2006robust} in large-scale problems does not seem to be the right practice, as RANSAC requires a  large number of iterations to achieve an acceptable fit. 

Curriculum Learning~\cite{bengio2009curriculum} begins learning a model from easy samples and gradually adapting the model to more complex ones, mimicking the cognitive process of humans. Ensemble Learning~\cite{dietterich2000ensemble} tries to improve the performance of machine learning algorithms by training different models and then to aggregate their predictions. Furthermore, distilling the knowledge learned from large deep learning models can be used to supervise a smaller model~\cite{hinton2015distilling}. Although Curriculum Learning, Ensemble Learning and distilling knowledge are notable methods, adopting them to work on problems with limited annotations, yet aside the unlabeled scenario, is far-from clear. 

\subsection{Deep Clustering}
Many research papers have explored clustering with deep neural networks. Deep Embedded clustering (DEC)~\cite{xie2016unsupervised} is one of the pioneers in this area, where the authors propose to pre-train a stacked auto-encoder (SAE)~\cite{bengio2007greedy} and fine-tune the encoder with a regularizer based on the student-t distribution  to achieve cluster-friendly embeddings. On the downside, DEC is sensitive to the network structure and initialization.
Various forms of  Generative Adversarial Network (GAN) are employed for clustering such as Info-GAN~\cite{chen2016infogan} and ClusterGAN~\cite{DBLP:journals/corr/abs-1809-03627}, both of which intend to enforce the discriminative feature in the latent space to simultaneously generate  and cluster images. 
The Deep Adaptive image Clustering (DAC)~\cite{chang2017deep} uses fully convolutional neural nets~\cite{springenberg2014striving} as initialization to perform self-supervised learning, and achieves  remarkable  results on various clustering benchmarks. However, sensitivity to the network structure seems again to be a concern for DAC.

In this paper, we formulate subspace clustering as a binary classification problem through collaborative learning of two modules, one for image classification and the other for subspace affinity learning. Instead of performing spectral clustering on the whole dataset, we train our model in a stochastic manner, leading to a scalable paradigm for subspace clustering. 

\section{Proposed Method}

To design a scalable SC algorithm, our idea is to identify whether a pair of points lies on the same subspace or not. Upon attaining such knowledge (for a large-enough set of pairs),  a deep model can optimize its weights to maximize such relationships (lying on  subspaces or not). 
This can be nicely cast as a binary classification problem. However, since ground-truth labels are not available to us, it is not obvious how such a  classifier should be built and trained.

In this work, we propose to make use of two confidence maps (see Fig.~\ref{structure} for a conceptual visualization) as a supervision signal for SC. To be more specific, we make use of the concept of self-expressiveness to identify positive pairs, \ie, pairs that lie on the same subspaces. To identify negative pairs, pairs that do not belong to same subspaces, we benefit from a negative confidence map. 
This, as we will show later, is due to the fact that the former can confidently mine positive pairs (with affinity close to 1) while the latter is good at localizing negative pairs (with affinity close to 0).
The two confidence maps, not only provide the supervision signal to optimize a deep model, but act collaboratively as partial supervisions for each other.

\subsection{Binary Classification}
\begin{figure}[!t]
\vskip 0.2in
\begin{center}
\centerline{\includegraphics[width=\columnwidth]{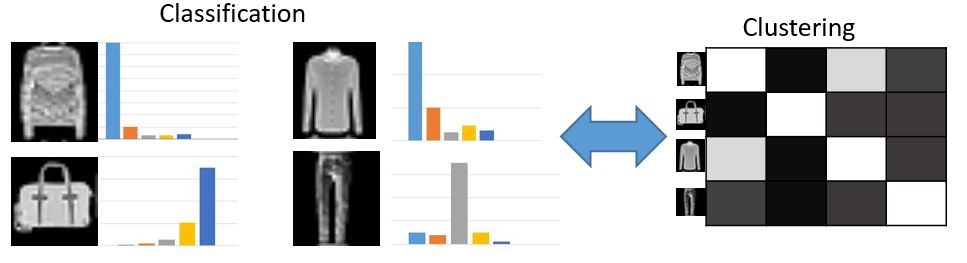}}
\caption{By normalizing the feature vectors after softmax function and computing their inner product, an affinity matrix can be generated to encode the clustering information. }
\label{classifier}
\end{center}
\vskip -0.2in
\end{figure}

Given a dataset with $n$ points $\mathcal { X } = \left\{ \mathbf { x } _ { i } \right\} _ { i = 1 } ^ { n }$ from $k$ clusters, we aim to train a classifier to predict class labels for data points without using the groundtruth labels. To this end, we propose to use a multi-class classifier which consists of a few convolutional layers (with non-linear rectifiers) and a softmax output layer. We then convert it to an affinity-based binary classifier by
\begin{equation}
    \mathbf{A}_{c}(i,j) =
     \mathbf{\nu}_i\mathbf{\nu}_j^T,
\end{equation}
where $\mathbf{\nu}_i\in \mathbb{R}^{k}$ is a $k$ dimensional prediction vector after $\ell_2$ normalization. Ideally, when $\mathbf{\nu}_i$ is one-hot, $\mathbf{A}_{c}$ is a binary matrix encoding the confidence of data points belonging to the same cluster. So if we supervise the classifier using $\mathbf{A}_{c}$, we will end up with a binary classification problem. Also note that $\mathbf{A}_{c}(i,j)$ can be interpreted as the cosine similarity between softmax prediction vectors of $\mathbf { x } _ { i }$ and $\mathbf { x } _ { j }$, which has been widely used in different contexts~\cite{nguyen2010cosine}. However, unlike the cosine similarity which lies in $[-1,1]$, $\mathbf{A}_{c}(i,j)$ lies within $[0,1]$, since the vectors are normalized by softmax and $\ell_2$ norm. We illustrate this in Fig.~\ref{classifier}.

\subsection{Self-Expressiveness Affinity}
Subspace self-expressiveness can be worded as: one data point $\mathbf{x}_i$ drawn from linear subspaces $ {\left\{ \mathcal { S } _ { i } \right\} _ { i = 1}^{k}}$ can be represented by a linear combination of other points from the same subspace. Stacking all the points into columns of a data matrix $\mathbf{X}$, the self-expressiveness can be simply described as $\mathbf{X} = \mathbf{X}\mathbf{C}$, where $\mathbf{C}$ is the coefficient matrix. 

It has been shown (\eg,~\cite{ji2014efficient}) that by minimizing certain norms of coefficient matrix $\mathbf{C}$, a block-diagonal structure (up to certain permutations) on $\mathbf{C}$ can be achieved. This translates into $c_{ji} \neq 0$ only if data points coming from the same subspace. Therefore, the loss function of learning the affinity matrix can be written as:
\begin{equation}\label{eq:self-expressive}
   \min\limits_{\mathbf{C}} \|{\mathbf{C} }\|_p  \quad
  \rm s.t. \quad {\bf X}  = {\bf X}{\bf C} ,\;  {\rm diag}({\bf C}) = {\bf 0} ,
\end{equation}
where $\|\cdot\|_p$ denotes a matrix norm. For example, Sparse Subspace Clustering (SSC)~\cite{elhamifar2009sparse} sticks to the $\ell_1$ norm, Low Rank Representation (LRR) models~\cite{liu2011latent, vidal2014low} pick the nuclear norm, and Efficient Dense Subspace Clustering~\cite{ji2014efficient} uses the $\ell_2$ norm. To handle data corruption, a relaxed version can be derived as:
\begin{equation}\label{eq:self-expressive_relax}
\begin{split}
  {\bf C}^* = &\argmin _ { \mathbf { C } } \| \mathbf { C } \| _ { p } + \frac { \lambda } { 2 } \| \mathbf { X } - \mathbf { X } \mathbf { C } \| _ { F } ^ { 2 } \quad \\
    & {\rm s.t.} \quad \operatorname {\rm diag } ( \mathbf { C } ) = \mathbf { 0 } .
  \end{split}
\end{equation}
Here, $\lambda$ is a weighting parameter balancing the regularization term and the data fidelity term.

To handle subspace non-linearity, one can employ convolutional auto-encoders to non-linearly map input data $\mathbf{X}$ to a latent space $\mathbf{Z}$, and transfer the self-expressiveness into a linear layer (without non-linear activation and bias parameters) named {\it self-expressive layer}~\cite{ji2017deep} (see the bottom part of Fig.~\ref{structure}). This enables us to learn the subspace affinity $\mathbf{A}_{s}$ in an end-to-end manner using the weight parameters ${\bf C}^*$ in the self-expressive layer:
\begin{equation}\label{eq:definition-As}
    \mathbf{A}_{s}(i,j) = \begin{cases}
    (|c^*_{ij}|+|c^*_{ji}|) / 2c_{\rm max} & \text{if $i\neq j$,} \\
    1 & \text{if $i=j$,}
    \end{cases}
\end{equation}
where $c_{\rm max}$ is the maximum absolute value of off-diagonal entries of the current row. Note that $\mathbf{A}_{s}(i,j)$ then lies within $[0,1]$.

\subsection{Collaborative Learning}
The purpose of collaborative learning is to find a principled way to exploit the advantages of different modules. The classification module and self-expressive module distill different information 
in the sense that the former tends to extract more abstract and discriminative features while the latter focuses more on capturing the pairwise correlation between data samples. From our previous discussion, ideally, the subspace affinity $\mathbf{A}_{s}(i,j)$ is nonzero only if $\mathbf{x}_i$ and $\mathbf{x}_j$ are from the same subspace, which means that $\mathbf{A}_{s}$ can be used to mine similar pairs (\ie, positive samples). On the other hand, if the classification affinity $\mathbf{A}_{c}(i,j)$ is close to zero, it indicates strongly  that $\mathbf{x}_i$ and $\mathbf{x}_j$ are dissimilar (\ie, negative sample). Therefore, we carefully design a mechanism to let both modules collaboratively supervise each other.

Given $\mathbf{A}_{c}$ and $\mathbf{A}_{s}$, we pick up the high-confidence affinities as supervision for training. We illustrate this process in Fig.~\ref{structure}. The ``positive confidence'' in Fig.~\ref{structure} denotes the ones from the same class, and the ``negative confidence'' represents the ones from different classes. 
As such, we select high affinities from $\mathbf{A}_{s}$ and small affinities from $\mathbf{A}_{c}$, and formulate the collaborative learning problem as:
\begin{equation}\label{Collaborative}
\begin{split}
& \min_{\mathbf{A}_{s},\mathbf{A}_{c}} \mathbf{\Omega}(\mathbf{A}_{s},\mathbf{A}_{c},l,u) =  \\
  L_{pos}&(\mathbf{A}_{s}, \mathbf{A}_{c},u) + \alpha L_{neg}(\mathbf{A}_{c},\mathbf{A}_{s},l), 
\end{split}
\end{equation}
where the $L_{pos}(\mathbf{A}_{s},\mathbf{A}_{c},u)$ and $L_{neg}(\mathbf{A}_{c},\mathbf{A}_{s},l)$ denote the cross-entropy function with sample selection process, which can be defined as follows:
\begin{equation}
\begin{split}\label{positive}
     L_{pos}(&\mathbf{A}_{s}, \mathbf{A}_{c},u) = H(\mathbf{M}_{s} || \mathbf{A}_{c} )  \\
&   \mathrm{s.t} \quad \mathbf{M}_{s} = \mathds{1}(\mathbf{A}_{s} > u),
\end{split}
\end{equation}
and 
\begin{equation}
\begin{split}\label{negative}
     L_{neg}(\mathbf{A}_{c}&, \mathbf{A}_{s},l ) = H(\mathbf{M}_{c} || (\mathbf{1} - \mathbf{A}_{s} ) )  \\
&  \mathrm{s.t} \quad \mathbf{M}_{c} = \mathds{1}(\mathbf{A}_{c} < l), 
\end{split}
\end{equation}
where $\mathds{1}(\cdot)$ is the indicator function returning $1$ or $0$, $\{u,l\}$ are thresholding parameters, and $H$ is the entropy function, defined as $H(\mathbf{p}||\mathbf{q}) = \sum_j p_j \log(q_j)$. 

Note that the cross-entropy loss is a non-symmetric metric function, where the former probability serves a supervisor to the latter. Therefore, in Eqn.~\eqref{positive}, the subspace affinity matrix $\mathbf{A}_{s}$ is used as the ``teacher'' to supervise the classification part (the ``student''). 
Conversely, in Eqn.~\eqref{negative}, the classification affinity matrix $\mathbf{A}_{c}$ works as the ``teacher'' to help the subspace affinity learning module to correct negative samples. However, to better facilitate gradient back-propagation between two modules, we can approximate indicator function by replacing $\mathbf{M}_{s}$ with $\mathbf{M}_{s} \mathbf{A}_{s}$ in Eqn.~\eqref{positive} and $\mathbf{M}_{c}$ with $ \mathbf{M}_{c} (1-\mathbf{A}_{c})$ in Eqn.~\eqref{negative}. The weight parameter $\alpha$ in Eqn.~\ref{Collaborative}, called collaboration rate, controls the contributions of $L_{pos}$ and $L_{neg}$. It can be set as the ratio of the number of positive confident pairs and the negative confident pairs, or slightly tuned for better performance.

\subsection{Loss Function}
After introducing all the building blocks of this work, we now explain how to jointly organize them in a network and train it with a carefully defined loss function. 
As shown in Fig.~\ref{structure}, our network is composed of four main parts: (i) a convolutional encoder that maps input data ${\bf X}$ to a latent representation ${\bf Z}$; (ii) a linear self-expressive layer which learns the subspace affinity through weights ${\bf C}$; (iii) a convolutional decoder that maps the data after self-expressive layer, \ie, ${\bf Z}{\bf C}$, back to the input space $\hat{\bf X}$; (iv) a multi-class classifier that outputs $k$ dimensional prediction vectors, with which a classification affinity matrix can be constructed.
Our loss function consists of two parts, \ie, collaborative learning loss and subspace learning loss, which can be written as:
\begin{equation}\label{eq:cost_function}
\begin{split}
    \mathds{L}(\mathbf{X};\Theta) & =  L_{sub}(\mathbf{X};\Theta,\mathbf{A}_{s}) \\
    & + \lambda_{cl} \mathbf{\Omega}(\mathbf{X},u,l;\Theta, \mathbf{A}_{s},\mathbf{A}_{c}),
\end{split}
\end{equation}
where $\Theta$ denotes the neural network parameters and $\lambda_{cl}$ a weight parameter for the collaborative learning loss.
The $L_{sub}(\mathbf{X};\Theta,\mathbf{A}_{s})$ is the loss to train the affinity matrix through self-expressive layer. Combining Eqn.~\eqref{eq:self-expressive_relax} and the reconstruction loss of the convolutional auto-encoder, we arrive at:
\begin{equation}
\begin{split}
\label{eq:loss_subs}
L_{sub}(\mathbf{X};\Theta,\mathbf{A}_{s}) & = \|\mathbf{C}\|_F^2 + \frac{\lambda_1}{2}\|{\bf Z} - {\bf Z}\mathbf{C}\|_F^2   \\
& + \frac{1}{2}\|{\bf X} - \hat{\bf X}\|_F^2 \quad \\
&{\rm s.t.}\quad  {\rm diag}(\mathbf{C}) = {\bf 0} ,
\end{split}
\end{equation}
where $\mathbf{A}_{s}$ is a function of ${\bf C}$ as defined in \eqref{eq:definition-As}.

After the training stage, we no longer need to run the decoder
 and self-expressive layer to infer the labels. We can directly infer the cluster labels through the classifier output $\mathbf{\nu}$:
\begin{equation}\label{eq:cluster}
    s_i = \argmax _h  \mathbf{\nu}_{ih},\;\; h = 1, \dots, k,
\end{equation}
where $s_i$ is the cluster label of image $\mathbf{x}_i$. 

\section{Optimization and Training}
\begin{algorithm}[tb]
   \caption{Neural Collaborative Subspace Clustering}\label{algorithm}
   \begin{algorithmic}
      \STATE   {\bfseries Input:} dataset $\mathcal { X } = \left\{ \mathbf{ x } _ { i } \right\} _ { i = 1 } ^ { N }$, number of clusters $k$, sample selection threshold $u$ and $l$ , learning rate of auto-encoder $\eta_{ae}$, and learning rate of other parts $\eta$ 
      \STATE {\bfseries Initialization:} Pre-train the Convolutional Auto-encoder by minimizing the reconstruction error.
      \REPEAT
      \STATE For every mini-batch data
      \STATE Train auto-encoder with {\it self-expressive layer} to minimize loss function in Eqn.~\eqref{eq:loss_subs} to update $\mathbf{A}_{s}$.
      \STATE Forward the batch data through the classifier to get $\mathbf{A}_{c}$.
      \STATE Do sample selection and collaborative learning through minimizing Eqn.~\eqref{Collaborative} to update the classifier.
      \STATE Jointly update all the parameters by minimizing Eqn.~\eqref{eq:cost_function}.
      \UNTIL{reach the maximum epochs}
      \STATE {\bfseries Output:} Get the cluster $s_i$ for all samples by Eqn.~\eqref{eq:cluster}
 \end{algorithmic}
\end{algorithm}
In this section, we provide more details  about how training will be done. Similarly to other auto-encoder based clustering methods, we pretrain the auto-encoder by minimizing the reconstruction error to get a good
initialization of latent space for subspace clustering.

According to~\cite{elhamifar2009sparse}, the solution to formulation~\eqref{eq:self-expressive} is guaranteed to have block-diagonal structure (up to certain permutations) under the assumption that the subspaces are independent. To account for this, we make sure that the dimensionality of the latent space (${\bf Z}$) is greater than (the subspace intrinsic dimension) $\times$ (number of clusters)~\footnote{Note that our algorithm does not require specifying subspace intrinsic dimensions explicitly. Empirically, we found a rough guess of the subspace intrinsic dimension would suffice, \eg, in most cases, we can set it to 9.}. 
In doing so, we make use of the stride convolution to down-sample the images while increasing the number of channels over layers to keep the latent space dimension large. 
Since we have pretrained the auto-encoder, we use a smaller learning rate in the auto-encoder when the collaborative learning is performed.
Furthermore, compared to DSC-Net or other spectral clustering based methods which require to perform sophisticated techniques to post process the affinity matrix, we only need to compute $\mathbf{A}_{s} = (|\mathbf{C}^*| + |\mathbf{C}^{*T}|)/2$ and normalize it (divided by the largest value in each row and assign $1.0$ to the diagonal entries) to ensure the subspace affinity matrix lie in the same range with the classification affinity matrix. 

We adopt a three-stage training strategy: first, we train the auto-encoder together with the self-expressive layer using the loss in~\eqref{eq:loss_subs} to update the subspace affinity $\mathbf{A}_{s}$; second, we train the classifier to minimize Eqn.~\eqref{Collaborative}; third, we jointly train the whole network to minimize the loss in~\eqref{eq:cost_function}. All these details are summarized in Algorithm~\ref{algorithm}.



\section{Experiments}
We implemented our framework with Tensorflow-1.6~\cite{abadi2016tensorflow} on a Nvidia TITAN X GPU. We mainly evaluate our method on three standard datasets, \ie, MNIST, Fashion-MNIST and the subset of Stanford Online Products dataset. All of these datasets are considered challenging for subspace clustering as it is hard to perform spectral clustering on datasets of this scale, and the linearity assumption is not valid.  The number of clusters $k$ is set to 10 as input to all competing algorithms. For all the experiments, we pre-train the convolutional auto-encoder for 60 epochs with a learning rate $1.0 \times 10^{-3}$ and use it as initialization, then decrease the learning rate to $1.0 \times 10^{-5}$ in training stage.

The hyper parameters in our loss function are easy to tune. $\lambda_1$ in Eqn.~\eqref{eq:loss_subs} controls  self-expressiveness, and it also affects the choice of  $u$ and $l$ in Eqn.~\eqref{eq:cost_function}. If $\lambda_1$ set larger, the coefficient in affinity matrix will be larger, and in that case the $l$ should be higher. The other parameter $\lambda_{cl}$ balances the cost of subspace clustering and collaborative learning, and we usually set it to keep these two terms in the same scale to treat them equally. 
We keep the $\lambda_1 = 10$ in all experiments, and slightly change the $l$ and $u$ for each dataset.


Our method is robust to different network design choices. We test different structures in our framework and get similar results on the same datasets. For MNIST, we use a three-layer convolutional encoder; for Fashion-MNIST and Stanford online Product, we use a deeper network consisting of three residual blocks~\cite{he2016deep}. We do not use batch normalization in our network because it will corrupt the subspace structure that we want to learn in latent space. We use the Rectified Linear Unit (ReLu) as the non-linear activation in our all experiments. 

Since there are no ground truth labels, we choose to use a larger batch size compared with supervised learning to make the training stable and robust. Specifically, we set the batch size to 5000, and use Adam~\cite{kingma2014adam}, an adaptive momentum based gradient descent method to minimize the loss for all our experiments. We set the learning rate to $1.0 \times 10^{-5}$  the auto-encoder and $1.0 \times 10^{-3}$ for other parts in all training stages. 

\textbf{Baseline Methods.} We use various clustering methods as our baseline methods including the classic clustering methods, subspace clustering methods, deep clustering methods, and GAN based methods. Specifically, we have the following baselines:
\begin{itemize}[noitemsep,topsep=0pt]
\item classic methods: $K$-Means~\cite{lloyd1982least} (KM), $K$-Means with our CAE-feature (CAE-KM) and SAE-feature (SAE-KM); 
\item subspace clustering algorithms: sparse subspace clustering (SSC)~\cite{elhamifar2013sparse}, Low Rank Representation (LRR)~\cite{liu2013robust}, Kernel Sparse Subspace Clustering (KSSC)~\cite{patel2014kernel}, Deep Subspace Clustering Network (DSC-Net)~\cite{ji2017deep}, and $k$-Subspace Clustering Network ($k$-SCN)~\cite{DBLP:journals/corr/abs-1811-01045}; 
\item deep clustering methods: Deep Embedded Clustering (DEC)~\cite{xie2016unsupervised}, Deep Clustering Network (DCN)~\cite{pmlr-v70-yang17b}, and Deep Adaptive image Clustering (DAC)~\cite{chang2017deep}; 
\item GAN based clustering methods: Info-GAN~\cite{chen2016infogan} and ClusterGAN~\cite{DBLP:journals/corr/abs-1809-03627}.
\end{itemize}

\textbf{Evaluation Metric.} For all quantitative evaluations, we make use of the unsupervised clustering accuracy rate, defined as
\begin{equation}
{\rm ACC }\;\% =\max_M \frac{\sum_{i=1}^n \mathds{1}(y_i = M(c_i))}{n}\times 100\%\;.
\end{equation}
where $y_i$ is the ground-truth label, $c_i$ is the subspace assignment produced by the algorithm, and $M$ ranges over all possible one-to-one mappings between subspaces and labels. The mappings can be efficiently computed by the Hungarian algorithm.
 We also use normalized mutual information (NMI) as the additional quantitative standard. NMI scales from 0 to 1, where a smaller value means less correlation between predict label and ground truth label. Another quantitative metric is the adjusted Rand index (ARI), which is scaled between -1 and 1. It computes a similarity between two clusters by considering all pairs of samples and counting pairs that are assigned in the same or different clusters in ground truth and predicted clusters. The larger the ARI, the better the clustering performance. 

\subsection{MNIST}
MNIST consists of $70000$ hand-written digit images of size $28\times 28$.
Subspace non-linearity arises naturally for MNIST due to the variance of scale, thickness and orientation among all the images of each digit. We thus apply our method on this dataset to see how well it can handle this type of subspace non-linearity.

In this experiment, we use a three-layer convolutional auto-encoder and a self-expressive layer in between the auto-encoder for the subspace affinity learning module. The convolution kernel sizes are $5-3-3-3-3-5$ and channels are $10-20-30-30-20-10$. For the classification module, we connect three more convolutional layers after the encoder layers with kernel size 2, and one convolutional layer with kernel size 1 to output the feature vector. For the threshold parameters $u$ and $l$, we set them to $0.1$ and $0.7$ respectively in the first epoch of training, and increase $l$ to $0.9$ afterwards. Our algorithm took around 15 mins to finish training on a normal PC with one TITAN X GPU.

We report the clustering results of all competing methods in Table ~\ref{tab:MNIST_results_test}. Since spectral clustering based methods (\ie, SSC-CAE, LRR-CAE, KSSC-CAE, DSC-Net) can not apply on the whole dateset (due to memory and computation issue), we only use the 10000 samples to show how they perform. As shown in Table~\ref{tab:MNIST_results_test}, subspace algorithms do not perform very well even on 10000 samples. Although the DSC-Net is trapped by training the self-expressive layer, it outperforms other subspace clustering algorithm, which shows the potential of learning subspace structure using neural networks. On the other hand, DEC, DCN, $k$-SCN and our algorithm are all based on auto-encoder, which learn embeddings with different metrics to help clustering. However, our classification module boost our performance through making the latent space of auto-encoder more discriminative. Therefore, our algorithm incorporates the advantage of different classes, \eg, self-expressivess, nonlinear mapping and discriminativeness, and achieves the best results among all the algorithms thanks to the collaborative learning paradigm.


\begin{table}[ht]
\centering
\begin{tabular}{l|l|l|l|}
 & ACC(\%) & NMI(\%) & ARI(\%) \\
 \hline
CAE-KM &51.00 & 44.87 &33.52  \\ 
SAE-KM & 81.29& 73.78 & 67.00  \\ 
KM & 53.00 & 50.00  & 37.00  \\    \hline
DEC &84.30 & 80.00 & 75.00 \\ 
DCN & 83.31 &80.86 &74.87 \\ \hline
SSC-CAE  & 43.03& 56.81    &  28.58  \\ 
LRR-CAE& 55.18 &    66.54& 40.57     \\
KSSC-CAE&  58.48&   67.74&   49.38   \\ 
DSC-Net  & 65.92&   73.00 &    57.09   \\ 
$k$-SCN& 87.14 &  78.15 & 75.81 \\ \hline
 Ours & \textbf{94.09}  & \textbf{86.12}  &\textbf{87.52} \\\hline
\end{tabular}
\caption{Clustering results of different methods on MNIST. For all quantitative metrics, the larger the better. The best results are shown in bold.}\label{tab:MNIST_results_test}
\end{table}


\subsection{Fashion-MNIST}
\begin{figure}[!t]
\begin{center}
\centerline{\includegraphics[width=1.0\columnwidth]{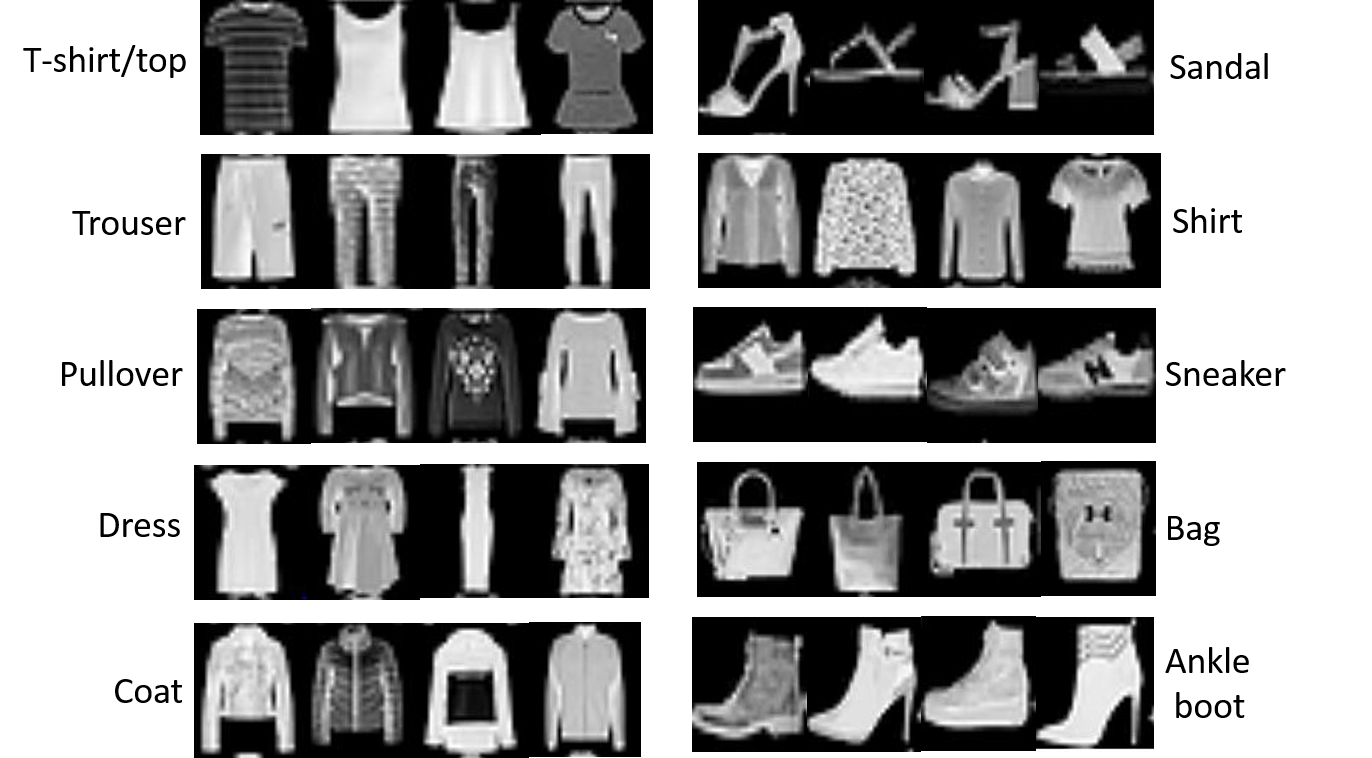}}
\caption{The data samples of the Fashion-Mnist Dataset }\label{samples_fashion}
\end{center}
\vskip -0.2in
\end{figure}

Same as in MINIST, Fashion-MNIST also has $70000$ images of size $28\times 28$. It consists of various types of fashion products. Unlike MNIST, every class in Fashion-MNIST has different styles with different gender groups (\eg, men, women, kids and neutral). As shown in Fig.~\ref{samples_fashion}, the high similarity between several classes (such as \{ Pullover, Coat, Shirt\}, \{ T-shirt, Dress \}) makes the clustering more difficult. Compared to MNIST, the Fashion-MNIST clearly poses more challenges for unsupervised clustering.

On Fashion-MNIST, we employ a network structure with one convolutional layer and three following residual blocks without batch normalization in the encoder, and with a symmetric structure in the decoder.  As the complexity of dataset increases, we also raise the dimensionality of ambient space to better suit self-expressiveness, and increase capacity for the classification module. For all convolutional layers, we keep kernel size as 3 and set the number of channels to 10-20-30-40. 

\begin{figure}[ht]
\begin{center}
\centerline{\includegraphics[width=0.5\columnwidth]{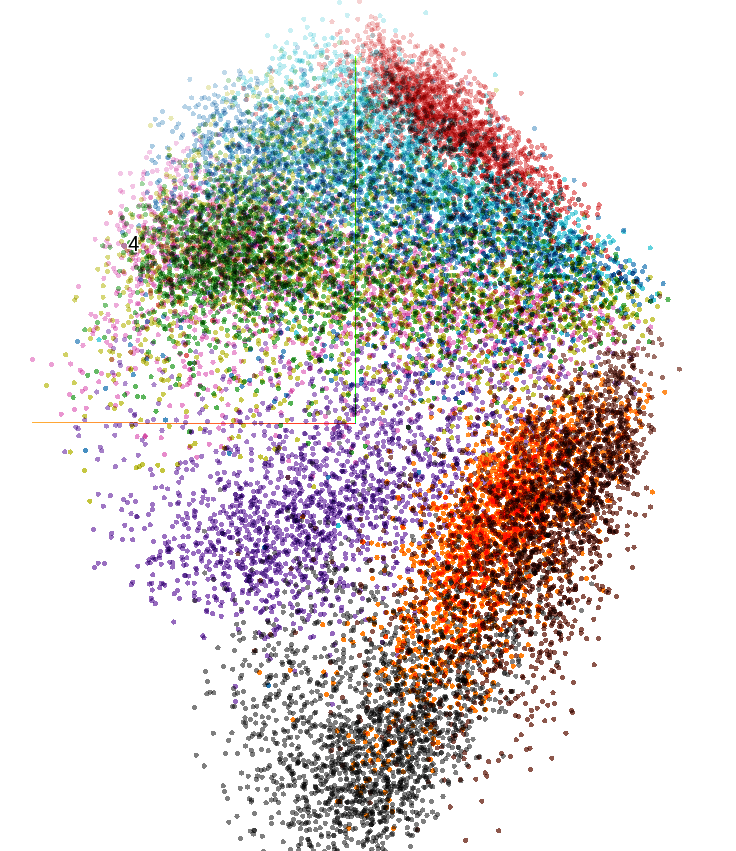}}
\caption{The visualization of our latent space through dimension reduction by PCA.}\label{fashion_visualization}
\end{center}
\vskip -0.2in
\end{figure}

We report the clustering results of all methods in Table~\ref{tab:Fashon-MNIST_results}, where we can clearly see that our framework outperforms all the baselines by a large margin including the the best-performing baseline $k$-SCN. Specifically, our method improves over the second one by $8.36\%$, $4.2 \%$ and $9\%$ in terms of accuracy, NMI and ARI. We can clearly observe from Fig.~\ref{fashion_visualization} that the latent space of our framework, which is collaboratively learned by subspace and classification modules, has strong subspace structure and also keeps each subspace discriminative. For subspace clustering methods we follow the way as on MNIST to use only 10000 samples. DSC-Net does not drop a lot while the performance of other subspace clustering algorithms decline sharply campared with their performance on MNIST. Since the code of ClusterGAN is not available currently, we can only provide results from their paper (without reporting ARI).

\begin{table}[ht]
\centering
\begin{tabular}{l|l|l|l|}
 & ACC(\%) & NMI(\%) & ARI(\%) \\
 \hline
SAE-KM & 54.35& 58.53 & 41.86  \\
CAE-KM & 39.84 & 39.80 & 25.93  \\ 
KM & 47.58 & 51.24  & 34.86  \\ \hline
DEC & 59.00 & 60.10 &  44.60 \\ 
DCN & 58.67 & 59.4& 43.04 \\  
DAC &   61.50  &  63.20    &  50.20  \\ 
ClusterGAN  &    63.00 &  64.00 & -   \\ 
InfoGAN &   61.00  &    59.00 & 44.20 \\  \hline
SSC-CAE & 35.87  & 18.10 &  13.46  \\
LRR-CAE & 34.48  & 25.41 &  10.33  \\
KSSC-CAE &  38.17    &   19.73 & 14.74  \\
DSC-Net &   60.62    & 61.71    & 48.20 \\ 
$k$-SCN& 63.78 &  62.04 & 48.04 \\ \hline
 Ours & \textbf{72.14}  & \textbf{68.60}  &\textbf{59.17} \\ \hline
\end{tabular}
\caption{Clustering results of different methods on Fashion-MNIST. For all quantity metrics, the larger the better. The best results are shown in bold.}
\label{tab:Fashon-MNIST_results}
\end{table}

\subsection{Stanford Online Products}

\begin{figure}[!t]
\vskip 0.2in
\begin{center}
\centerline{\includegraphics[width=1.1\columnwidth]{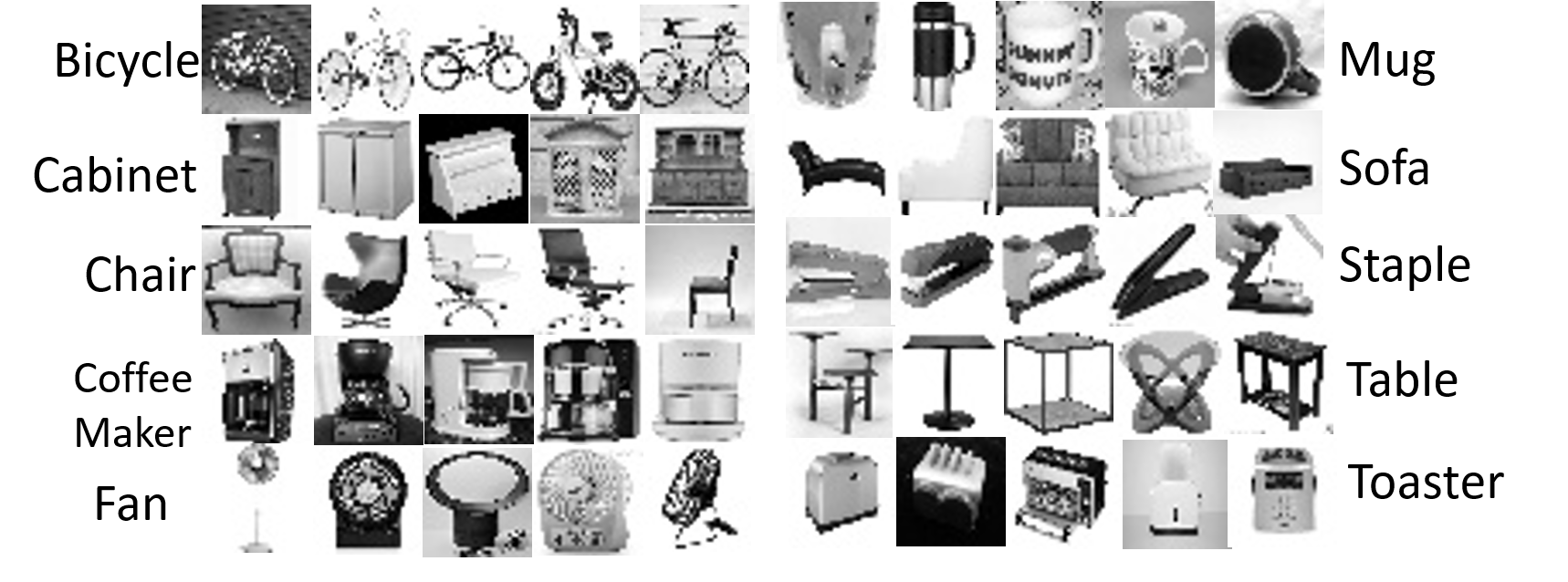}}
\caption{The data samples of the Stanford Online Products Dataset }\label{samples_stanford}
\end{center}
\vskip -0.2in
\end{figure}

The Stanford Online Products dataset is designed for supervised metric learning, and it is thus considered to be difficult for unsupervised clustering. Compared to the previous two, the challenging aspects of this dataset include: (i) the product images contain various backgrounds, from pure white to real world environments; (ii) each product has different shapes, colors, scales and view angles; (iii) products across different classes may look similar to each other. To create a manageable dataset for clustering, we manually pick 10 classes out of 12 classes, with around 1000 images per class (10056 images in total), and then re-scale them to $32 \times 32$ gray images, as shown Fig.~\ref{samples_stanford}.  

Our networks for this dataset start from one layer convolutional kernel with 10 channels, and follow with three pre-activation residual blocks without batch normalization, which have 20, 30 and 10 channels respectively. 

Table~\ref{stanford_table} shows the performance of all algorithms on this dataset. Due to the high difficulty of this dataset, most deep learning based methods fail to generate reasonable results. For example, DEC and DCN perform even worse than their initialization, and DAC can not self-supervise their model to achieve a better result. Similarly, infoGAN  also fails to find enough clustering pattern. 
In contrast, our algorithm achieves better results compared to other algorithms, especially the deep learning based algorithms. Our algorithm along with KSSC and DSC-Net achieve top results, due to the handling of non-linearity. Constrained by the size of dataset, our algorithm does not greatly surpass the KSSC and DSC-Net. We can easily observe that subspace based clustering algorithms perform better than clustering methods. This illustrates how effective the underlying subspace assumption is in high dimension data space, and it should be considered to be a general tool to help clustering in large scale datasets. 

In summary, compared to other deep learning methods, our framework is not sensitive to the architecture of neural networks, as long as the dimensionality meets the requirement of subspace self-expressiveness. 
Furthermore, the two modules in our network progressively improve the performance in a collaborative way, which is both effective and efficient.

\begin{table}[!t]
\centering
\begin{tabular}{l|l|l|l|l|}
 & ACC (\%) & NMI (\%) & ARI (\%)\\ \hline
DEC & 22.89 & 12.10 & 3.62 \\ 
DCN & 21.30 & 8.40 &   3.14  \\ 
DAC  &  23.10 &  9.80    & 6.15 \\
InfoGAN & 19.76   &   8.15  & 3.79  \\ \hline
SSC-CAE   &12.66  &   0.73  & 0.19 \\ 
LRR-CAE   & 22.35  & \textbf{17.36} & 4.04   \\ 
KSSC-CAE & 26.84 &   15.17      & 7.48 \\ 
DSC-Net &   26.87 &   14.56     & \textbf{8.75}   \\   
$k$-SCN& 22.91 &  16.57  &   7.27\\ \hline
Ours & \textbf{27.5}  & 13.78 & 7.69  \\  \hline
\end{tabular}
\caption{The clustering results of different algorithms on subset of Stanford Online Products. The best results are in bold.}\label{stanford_table}
\end{table}

%


\section{Conclusion}


In this work, we have introduced a novel learning paradigm, dubbed collaborative learning, for unsupervised subspace clustering. To this end, we have analyzed the complementary property of the classifier-induced affinities and the subspace-based affinities, and have further proposed a collaborative learning framework to train the network. Our network can be trained in a batch-by-batch manner and can directly predict the clustering labels (once trained) without performing spectral clustering. The experiments in our paper have shown that the proposed method outperforms the-state-of-art algorithms by a large margin on image clustering tasks, which validates the effectiveness of our framework.

\nocite{langley00}

\bibliography{deepsubspace.bib}
\bibliographystyle{icml2019}


\end{document}